\documentclass[sigconf]{acmart}

\usepackage{multirow}
\usepackage{graphicx}
\usepackage{spverbatim}

\AtBeginDocument{%
  }

\renewcommand\footnotetextcopyrightpermission[1]{} %

\begin{document}
\fancyhf{} %
\pagestyle{empty} %

\title{You Can Generate It Again: Data-to-Text Generation with Verification and Correction Prompting}

\author{Xuan Ren\quad Zeyu Zhang\quad Lingqiao Liu\\
\vspace{0.2cm}
AIML, University of Adelaide}

\begin{abstract}
Small language models like T5 excel in generating high-quality text for data-to-text tasks, offering adaptability and cost-efficiency compared to Large Language Models (LLMs). However, they frequently miss keywords, which is considered one of the most severe and common errors in this task.
In this work, we explore the potential of using feedback systems to enhance semantic fidelity in smaller language models for data-to-text generation tasks, through our Verification and Correction Prompting (VCP) approach.
In the inference stage, our approach involves a multi-step process, including generation, verification, and regeneration stages. During the verification stage, we implement a simple rule to check for the presence of every keyword in the prediction. Recognizing that this rule can be inaccurate, we have developed a carefully designed training procedure, which enabling the model to incorporate feedback from the error-correcting prompt effectively, despite its potential inaccuracies.
The VCP approach effectively reduces the Semantic Error Rate (SER) while maintaining the text's quality.
\end{abstract}



\maketitle

\section{Introduction}

\begin{figure}[t]
   \centering
    \includegraphics[width=0.85\columnwidth]{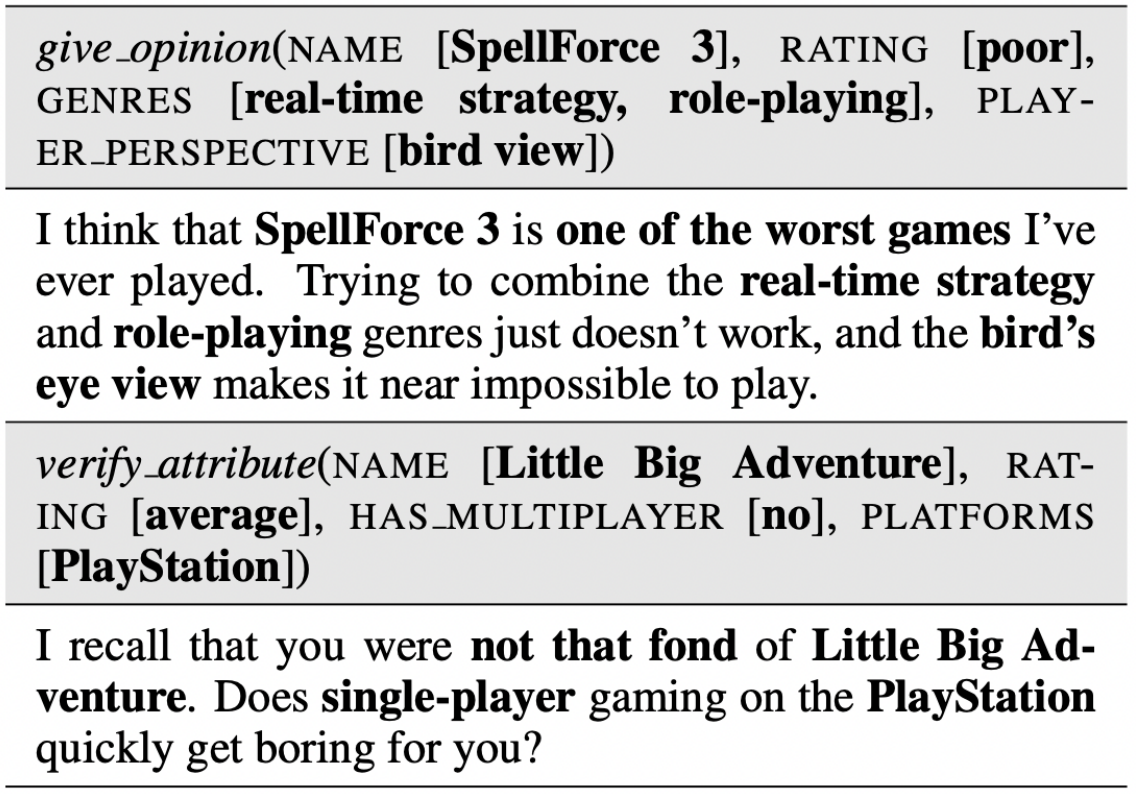}
    \vspace{-0.3cm}
    \caption{Data-to-text examples from the ViGGO dataset \cite{juraska-etal-2019-viggo}.} 
    \label{fig:Figure1}
    \vspace{-0.55cm}
\end{figure}

Data-to-text generation aims to convert structured data into coherent, human-readable text as shown in \hyperref[fig:Figure1]{Figure~\ref{fig:Figure1}}. It has a broad range of practical applications, including report generation, automated journalism, data visualization, and dialogue systems, or it can be used as intermediate steps in large projects. In these applications, the input can consist of various types of data, such as tables, graphs, or raw data. It is worth noting that data-to-text generation is a controlled form of text generation, where the output must be coherent with the input and maintain semantic accuracy.

Fine-tuning pre-trained small models, such as T5 \citep{10.5555/3455716.3455856}, which are more efficient compared to LLMs, is often sufficient for many data-to-text generation tasks, as these tasks do not require strong reasoning skills. The main challenge lies in ensuring adherence to instructions and accurately replicating specific text styles. One of the most severe and frequent problems is omissions, the absence of crucial keywords \citep{yin-wan-2022-seq2seq}. For instance, in the first example from \hyperref[fig:Figure1]{Figure~\ref{fig:Figure1}}, if the prediction omits the name 'SpellForce 3', then the prediction has one slot error. In this paper, we introduce the 'slot error rate (SER),' which quantifies the rate of missing keywords.

Several research works aim to reduce the SER, including the copy mechanism \citep{rebuffel2019hierarchical, puduppully2019data}, template-based generation \citep{kale2020template, mehta-etal-2022-improving}, planning-then-generate \citep{xu-etal-2021-agggen, su2021plan, kasner-dusek-2022-neural}, and post-editing \citep{jolly2022search, balachandran-etal-2022-correcting}. These methods often rely on strict rules and can effectively reduce SER but may sacrifice text fluency. Techniques such as those by \citep{juraska-walker-2021-attention, seifossadat2023improving} guide attention behavior, leading the model to make more accurate generations. Such methods are flexible, thus reducing SER while maintaining text fluency.

Regeneration according to feedback \citep{madaan2023self, xue2023rcot} has recently gained popularity, predominantly in LLMs. This approach requires an accurate feedback system to generate natural language feedback prompts. However, smaller language models, focused on efficiency, may not interpret these feedback prompts accurately and often lack a precise verifier. Employing an accurate verifier, such as an LLM or meticulously handcrafted rules, would contradict the original goal of prioritizing efficiency. We aim to explore the feasibility of using a feedback system with a trivial verifier to enhance the semantic accuracy of a smaller model in data-to-text generation.

\definecolor{darkgreen}{rgb}{0.0, 0.5, 0.0}

\begin{table*}
\caption{The inference process comprises three steps: 1. We utilize the fine-tuned T5 model to generate an initial prediction. 2. The slot error checker is then deployed to ascertain the presence of any slot errors. If such errors are detected, we label the error-correcting prompts to highlight the location of potential slot errors. 3. Lastly, we reintroduce the prompted input to the fine-tuned T5 model for regeneration. The tokens(error-correcting prompts) have the capacity to alter the regenerated outputs, ensuring the inclusion of previously missed slots.}
\vspace{-0.3cm}
\centering
\begin{tabular}{|p{0.25\textwidth}|p{0.7\textwidth}|}
\multicolumn{2}{l}{\textbf{Step1: Initial Prediction}} \\ \hline
Input & recommand(name(Tom Clancy{]}, release\_year{[}1999{]}, has\_linux\_release{[}yes{]}) \\ \hline
T5 predictions & Since you’re into Linux games, you heard of Tom Clancy? \\ \hline 
\multicolumn{2}{l}{\textbf{Step2: Verification}} \\ \hline
\multirow{3}{*}{Find slot errors} & 1999 \textcolor{red}{$\times$} \\ 
                                  & Tom Clancy \textcolor{darkgreen}{\checkmark }\\ 
                                  & Linux \textcolor{darkgreen}{\checkmark} \\ \hline
Label missing slots & recommand(release\_year{[}\textcolor{red}{\textless{}a1\textgreater \textless{}token2\textgreater \textless{}token3}\textgreater 1999{]}, name(Tom Clancy{]}, has\_linux\_release{[}no{]}) \\ \hline  
\multicolumn{2}{l}{\textbf{Step3: Regeneration}} \\ \hline
Prompted Input & recommand(release\_year{[}\textcolor{red}{\textless{}token1\textgreater \textless{}token2\textgreater \textless{}token3\textgreater{}}1999{]}, name(Tom Clancy{]}, has\_linux\_release{[}no{]}) \\ \hline
Send to T5 to generate prediction & Since you’re into Linux games, have you heard of Tom Clancy which is released in 1999? \\ \hline
\end{tabular}
\end{table*}

\begin{figure*}[t]
    \centering
    \includegraphics[width=\textwidth]{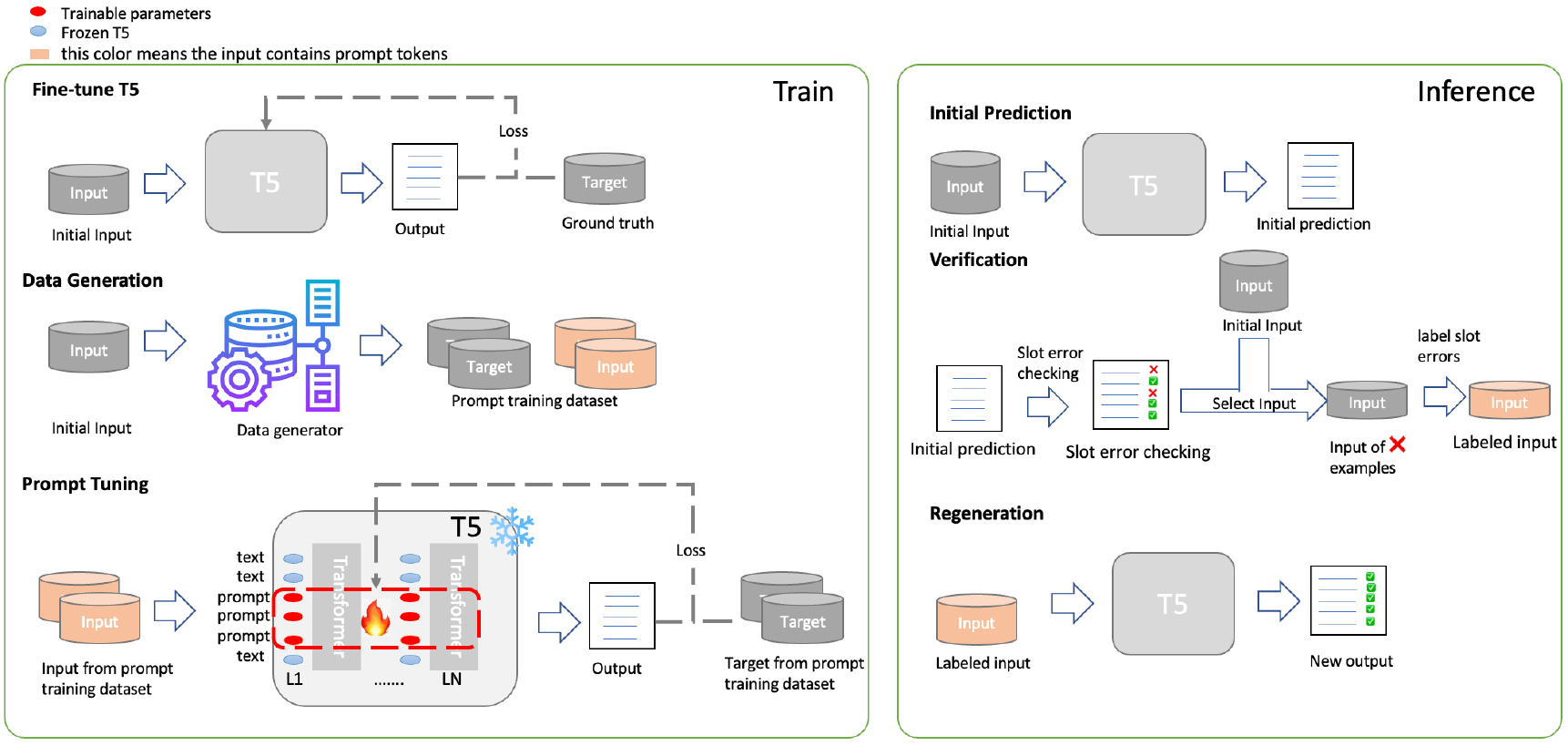}
    \caption{The workflow is comprised of train and inference section. When training, we first fine tune a T5 model, then use data generator to generate prompt training dataset. Lastly use prompt tuning to teach error-correcting prompts how to improve the semantic coverage in T5's prediction. The inference section illustrates the overview of the initial prediction, verification, and regeneration process. Please refer to \hyperref[tab:1]{Table 1} for more detailed insights.}
    \label{fig:Figure2} 
    \vspace{-0.2cm}
\end{figure*}

Inspired by previous works, we propose a prompt-based, test-time correction pipeline, VCP, designed to encourage the model to include missed slots identified by a slot error checker while maintaining text quality. The inference process is illustrated in \hyperref[fig:Figure2]{Figure 2}. Initially, the fine-tuned T5 model receives the input slots and generates initial predictions. The slot error checker then verifies whether any slots are missing from the output. If a slot is missing, we label the corresponding error-correcting prompts along with the missed input value. During the regeneration process, these error-correcting prompts guide the fine-tuned T5 model to include the omitted slot value in its subsequent prediction. An example is shown in \hyperref[tab:1]{Table 1}.

To enable the aforementioned error-correcting regeneration process, it is necessary to train error-correcting prompts that encourage the model to include slots it previously missed. Since the slot error checker relies on trivial rules, the trained prompts must guide the T5 model in such a way that it does not alter its prediction when the prompt is mislabeled. The training process for these prompts is outlined in the training section of \hyperref[fig:Figure2]{Figure 2}. Specifically, in the Data Generation process, a data generator is used to create unseen prompted inputs along with their corresponding ground truths. This serves to construct both the prompt initialization training dataset and the prompt training dataset. During training, we first train the prompt initialization and then fine-tune the prompt embedding based on the initialized prompt. These error-correcting prompts are designed to direct the fine-tuned T5 model to include the labeled slot values it previously missed in its predictions. During training, the model and the prompts are exposed to scenarios where slots are mislabeled by error-correcting prompts, teaching them to disregard inaccurate labels. Our method achieves a lower SER while maintaining competitive text fluency compared to other methods.

\section{Related Works and Comparative Analysis}
\label{sec:related works}

In data-to-text generation tasks, slot error, the omission of essential keywords, is a significant and frequent issue, as identified by \citep{yin-wan-2022-seq2seq}. To address this, various strategies have been developed:

1. Strict Generation Processes: Methods such as copy mechanisms \citep{rebuffel2019hierarchical, puduppully2019data}, template-based generation \citep{kale2020template, mehta-etal-2022-improving}, and plan-then-generate approaches \citep{xu-etal-2021-agggen, su2021plan, kasner-dusek-2022-neural} ensure the inclusion of all keywords. These methods enforce a model's generation to strictly adhere to the input structure. While effective in minimizing slot errors, these methods can lead to less fluent text due to their rigidity.

2. Post-editing Approaches: \citep{jolly2022search} search for missing keywords and find the best position to insert the phrase containing these keywords. They only conducted experiments in a few-shot setting. \citep{balachandran-etal-2022-correcting} adversarially train an error correction network to correct factual errors in summarizing tasks. The error correction training dataset is constructed by replacing correct factual words with incorrect ones. In data-to-text generation, missing even one keyword can disrupt the entire sentence, thus this method cannot be directly applied to data-to-text generation.

3. Attention Behavior Guidance: \citep{juraska-walker-2021-attention} manually identified three attention patterns associated with semantic errors. They created a script to automatically adjust the beam search scores according to these three attention patterns during inference. By adding a dynamic memory module to the attention-based network, DM-NLG \citep{seifossadat2023improving} can store previously generated words, thus better guiding the generation process to include key information. 

Moreover, the use of feedback systems for better predictions has been explored \citep{madaan2023self, xue2023rcot, peng2023check, shridhar2023art, shridhar2023screws}, particularly in reasoning tasks. These systems, which often utilize a Large Language Model (LLM), prior knowledge bases, or specific rules for verification, align with our work in their inferential process. Our approach is distinct from previous feedback systems in that: 1. Our verification process is efficient and trivial. It does not require prior knowledge or a LLM. 2. Our soft prompt feedback can guide small models that are not capable of responding to natural language feedback. 3. The text generation quality remain high even if the feedback is inaccurate. 4. Our VCP can maintain text generation styles. Our method diverges from traditional post-editing by employing error-correcting prompts for complete output regeneration, rather than just tweaking initial outputs. This enables flexible restructuring or filling in of missing information, leading to smoother and more coherent text.

\section{Methodology}
\hyperref[fig:Figure2]{Figure 2} provides a comprehensive illustration of our method, detailing both the inference and training stages involved in the process.

\subsection{Inference}
An inference example is presented in \hyperref[tab:1]{Table 1}, encompassing three steps: initial generation, verification, and regeneration. Initially, we feed the testing input into the fine-tuned T5 model, which was trained with the data-to-text training dataset, to generate the initial prediction. During the verification step, the Slot Error Checker, employing exact match, examines whether any slots from the input are missing in the output, thereby identifying slot errors. If any potential errors are detected, we introduce error-correcting prompts adjacent to the positions of the unmentioned slots in the input. Lastly, in the regeneration step, the prompted inputs are fed back into the fine-tuned T5 model. These error-correcting prompts guide the T5 model to incorporate the previously omitted slots during the regeneration.

\subsection{Slot Error Checking} 
The Slot Error Checker verifies the presence of slot errors using exact match. For non-boolean slot values, it simply checks if all of the slot values from the input are mentioned in the output. An example is shown in Table 1, where '1999' from the input slot is missing in the initial prediction and thus is identified as an error in the verification step. For boolean slot value pairs which utilize \textit{yes} or \textit{no} as slot values, we do not check for the presence of \textit{yes} or \textit{no} in the prediction. Instead, we focus on whether the noun part of the slot names, identified by part-of-speech (POS) tagging—such as \textit{linux} from \textit{has\_linux\_released} or \textit{mac} from \textit{has\_mac\_released}, is mentioned in the predictions. We utilize the same slot error checker for ViGGO and E2E, steering clear of any domain-specific knowledge. It solely depends on exact matches, leading to a high incidence of falsely indicated errors. Importantly, our approach refrains from employing sophisticated rules for slot error checking due to the lack of generalizability of these engineering rules across different datasets. Nonetheless, the error-correcting prompt adeptly learns to navigate mislabeling scenarios, as detailed in the section on \hyperref[sec:pt]{Prompt Tuning}. For an in-depth explanation of the slot error checker, please refer to the Appendix~\hyperref[sec:slot_error_checker_explanation]{Detailed Explanation of the Slot Error Checking Process}

\subsection{Training}
\label{sec:trainprompt}
The training procedure, as illustrated in \hyperref[fig:Figure2]{Figure 2}, begins by fine-tuning a T5 model using the original training dataset. This step establishes a well-initialized base model. Subsequently, a prompt training dataset is created for learning error-correcting prompts. Finally, while keeping the fine-tuned T5 model frozen, the error-correcting prompts are trained on the prompt training dataset to enhance their ability to guide the language model in integrating slot values that were previously missed. For more details, see the \hyperref[pt]{Prompt Tuning} section.

\begin{figure}[t]
   \centering
    \includegraphics[width=\columnwidth]{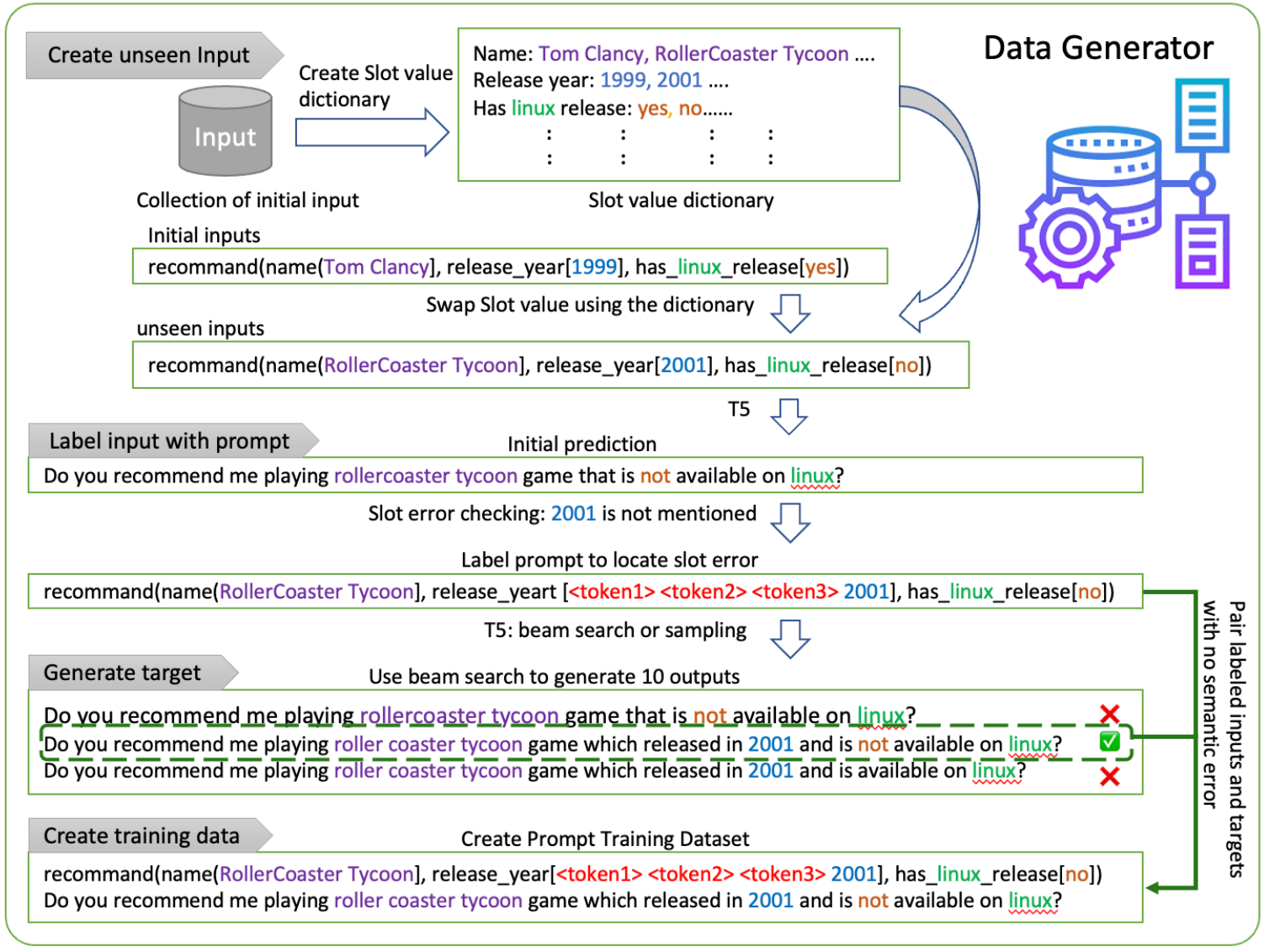}
    \caption{Generating prompt training datasets for training error-correcting prompts.}
    \label{fig:Figure3}
    \vspace{-0.5cm}
\end{figure}

\textbf{Generating Prompt Training Dataset}: Since the training data has already been exposed to the T5 model during the initial fine-tuning, we cannot use the same dataset to learn the error-correction prompts. Therefore, a new training set is generated for prompt learning, as depicted in \hyperref[fig:Figure3]{Figure 3}. Specifically, the data generation process (data generator) creates input-output pairs. The candidate input is generated by replacing the slot values in an input from the original training set with other values randomly sampled from the possible values for each slot. For example, "\textit{recommend(name[Tom Clancy], release\_year[1999], has\_linux\_release[yes])}" comprises the intention "\textit{recommend}", slot names "\textit{release\_year}, \textit{name}, and \textit{has\_linux\_release}", and slot values "\textit{1999}, \textit{Tom Clancy}, and \textit{yes}." We use a slot value dictionary, created from grouping all unique values corresponding to the same slot name in the training set, for this replacement. After slot value replacement, we could generate an unseen input like "\textit{recommend(name[RollerCoaster Tycoon], release\_year[2001], has\_linux\_release[no])}." These candidate inputs are then fed into the fine-tuned T5 model to generate initial predictions. The Slot Error Checker is applied to identify the parts of the inputs with slot errors, which are subsequently marked with an error-correcting prompt.

Then, ground-truth output for these prompted unseen inputs needs to be generated. We feed the gathered unseen inputs into the fine-tuned T5 model, employing beam search to generate 10 predictions, with the beam size set to 10. These predictions are then subjected to Slot error checking. The prediction identified as being free of slot errors is selected as the ground truth. In scenarios where multiple outputs from beam search are identified as free of slot errors, the output with the highest probability, as determined by the beam search, is chosen as the ground truth. The training dataset created contains numerous mislabeled examples, as the slot error checker lacks accuracy.

\textbf{Prompt Tuning}\label{pt}. Once sufficient input-output pairs are generated, we fine-tune the error-correcting prompts while keeping the T5 model fixed. The training process is designed to learn error-correcting prompts that guide the T5 model to produce outputs without missing slot values. As the prompt training dataset contains examples that are correctly or wrongly labeled, the prompts learn how to handle these situations during training. For instance, the slot 'RATING [poor]' is tagged with an error-correcting prompt because it does not align with the reference 'one of the worst games.' However, these error-correcting prompts do not affect the prediction during regeneration. This characteristic makes our method more robust compared to previous post-editing methods that rely on strict rules. For a detailed numerical analysis, refering to  Appendix~\hyperref[sec:error_correcting_prompts_robustness]{Robustness to Slot Checking Errors}.

In our design, we train deep prompts (P-Tuning v2 \citep{liu-etal-2022-p}). Specifically, we use 3 error-correcting prompts for T5-base and 6 error-correcting prompts for T5-small. The trainable prompts are added to each layer in T5, encompassing the word embeddings of the error-correcting prompts and the key-value embeddings in every layer.

\textbf{Prompt Initialization} 
In addition to the workflow shown in \hyperref[fig:Figure2]{Figure 2}, our ablation study finds the advantages of introducing a prompt initialization phase. This phase trains a robust initial embedding, ensuring that text generation quality is unaffected by prompt insertion. More details are described in the appendix.

\section{Experiment}
We compare our VCP method with various other approaches within the E2E \citep{novikova-etal-2017-e2e} and ViGGO datasets \citep{juraska-etal-2019-viggo}. Our primary comparison targets the two methods known for guiding attention behavior, SEA-GUIDE \citep{juraska-walker-2021-attention} and DM-NLG \citep{seifossadat2023improving}, due to their superior performance. Additional methods included in our analysis are K\&M \citep{kedzie-mckeown-2020-controllable}, which leverages data augmentation; DT \citep{harkous-etal-2020-text}, employing a generation-reranking approach; and the post-editing technique, T5 S\&L \citep{jolly2022search}, which is evaluated in a few-shot setting. The baseline method for the ViGGO dataset is S2S \citep{juraska-etal-2019-viggo}.

\subsection{Dataset and Evaluation Metrics} 
The experiments are conducted on the E2E and ViGGO datasets. The E2E dataset \citep{novikova-etal-2017-e2e}, specifically designed for the restaurant domain, offers a data-driven approach for end-to-end natural language generation system training. The ViGGO dataset \citep{juraska-etal-2019-viggo} targets open-domain dialogue systems in video game topics, covering 9 generalizable and conversational dialogue act types.

Our system's performance is assessed using a comprehensive set of metrics. For non-semantic error evaluation, we use BLEU \citep{papineni2002bleu}, METEOR \citep{lavie-agarwal-2007-meteor}, ROUGE \citep{lin-2004-rouge}, CIDEr \citep{vedantam2015cider}, which are assessed using the E2E evaluation script\footnote{\url{https://github.com/tuetschek/e2e-metrics}}. For semantic error evaluation, we use the SER which measures the error rate of slot values in the generated text. We apply the same auto slot evaluation script used in the SEA-GUIDE project\footnote{\url{https://github.com/jjuraska/data2text-nlg}}, encompassing hundreds of evaluation rules. They conducted a human evaluation of this script, demonstrating that the SER evaluation script has a 97.7\% accuracy rate on the ViGGO dataset and 100\% accuracy on the E2E dataset.

The SER is calculated by dividing the number of slot errors by the total number of slots. For instance, if a sentence contains 1 slot error but has a total of 8 slots, the SER for this incorrect sentence would be 12.5\% rather than 100\%. People unfamiliar with SER might underestimate the severity of a low SER rate. Thus, we introduce the Sentence-Level Semantic Error Rate (SLSER), which counts the percentage of sentences with semantic errors. This analysis was conducted manually to highlight the severity of the slot errors. We primarily use SER to measure slot errors, as this is the common metric used in other research.

\subsection{Setup}
We observed a substantial variance in the SER of the T5 baseline, SEA-GUIDE, and our method when applied to the ViGGO dataset. To ensure a fair comparison, we trained 5 instances each of the T5-small and T5-base models, each for 20 epochs. We also ran SEA-GUIDE and our VCP 5 times for T5-base and T5-small, calculating the mean and variance. All optimized models were selected based on validation loss. In the VCP project, we used a batch size of 10 and a maximum sentence length of 300 tokens. All tests were conducted on a single RTX 3090 GPU, with a linear learning rate scheduler. More comprehensive information regarding the prompt tuning hyperparameters is provided in the Appendix. For the E2E datasets, we followed the same procedure. The experimental results for other methods, where standard deviations are not reported, are as presented in Tables 2 and 3, and have been taken directly from the original papers.

\subsection{Performance Comparison}
\label{sec:performance comparison}

\begin{table}[t]
\caption{Comparison of VCP to other methods and T5 baseline on the ViGGO. SLSER represents the percentage of sentences containing a slot error.}
\vspace{-0.3cm}
\centering
\resizebox{\columnwidth}{!}{
\begin{tabular}{lllllll}
\hline
Model                                  & BLEU            & MET.          & ROUGE & CIDEr & SER ↓ & SLSER  \\ \hline
T5-small$_{\text{beam search}}$ baseline  & $53.2\pm 0.54$  & 0.392         & 0.637 & 2.652 & $0.89\pm 0.096\%$  & 3\%\\
T5-base$_{\text{beam search}}$ baseline   & $53.1\pm 0.28$  & 0.393         & 0.635 & 2.655 & $0.60\pm 0.13\%$ & 2\% \\ 
S2S                                    & 51.9            & 0.388         & 0.631 & 2.531 & 2.55\%  & - \\ 
DT                                     & 53.6            & 0.394         & 0.640 & 2.700 & 1.68\%  & - \\ 
K\&M                                   & 48.5            & 0.380         & 0.592 & 2.454 & \textit{0.46\%} & - \\ 
SEA-GUIDE$_{\text{T5-small}}$             & $53.2\pm 0.53$  & 0.392         & 0.637 & 2.693 & $0.7\pm 0.097\%$ & 2.0\% \\
SEA-GUIDE$_{\text{T5-base}}$              & $53.2\pm 0.30$  & 0.393         & 0.635 & 2.658 & $0.51\pm 0.10\%$ & 1.7\% \\
VCP$_{\text{T5-small}}$                  & $52.6\pm 0.51$  & 0.392         & 0.632 & 2.628 & $\textbf{0.41}\pm 0.085\%$ & 1.4\%  \\  
VCP$_{\text{T5-base}}$                   & $52.4\pm 0.19$  & 0.391         & 0.627 & 2.620 & $\textbf{0.33}\pm 0.19\%$ & 1.2\% \\ \hline
\end{tabular}
}
\label{tab:2}
\end{table}

\begin{table}[t]
\caption{Comparison of our method VCP to other methods on the E2E dataset. SLSER represents the percentage of sentences containing a slot error.}
\vspace{-0.3cm}
\centering
\resizebox{\columnwidth}{!}{
\begin{tabular}{llllll}
\hline
Model                                      & BLEU  & MET.  & ROUGE & SER ↓   & SLSER \\ \hline
T5 S\&L w/ SER p:420, u:1680                & 60.9 & 0.437 & 0.668  & 0.8\% & - \\
VCP$_{\text{T5-base}}$  p:420, u:1680      & 60.5 & 0.438 & 0.667 & 0.55\% & - \\ \hline
T5-small$_{\text{greedy search}}$ baseline & 67.0 & 0.454 & 0.692  & 1.60\% & 9.9\%\\
T5-small$_{\text{beam search}}$ baseline   & 66.7 & 0.453 & 0.694  & 2.85\% & 11.6\%\\
T5-base$_{\text{greedy search}}$ baseline  & 66.8 & 0.459 & 2.282  & 1.85\% & -\\
T5-base$_{\text{beam search}}$ baseline    & 66.7 & 0.453 & 0.697  & 3.94\% & - \\ \hline
S2S                                         & 66.2 & 0.445 & 0.677  & 0.91\% & - \\
K\&M                                        & 66.3 & 0.453 & 0.693  & \textbf{0} & - \\
SEA-GUIDE$_{\text{T5-small}}$               & 67.5 & 0.453 & 0.690  & 0.04\% & 0.25\% \\
SEA-GUIDE$_{\text{T5-base}}$                & 68.2 & 0.454 & 0.691  & 0.05\% & 0.32\% \\
DM-NLG no postprocess                       & 66.7 & 0.456 & 0.691  & 0.03\% & - \\
DM-NLG postprocess: GPT-2                   & 68.6 & 0.482 & 0.713  & 0.03\% & - \\
VCP$_{\text{T5-small}}$                     & $67.0\pm 0.18$ & 0.451 & 0.690 & \textbf{0.002\%} & 0.015\% \\ \hline

\end{tabular}
}
\label{tab:3}
\vspace{-0.3cm}
\end{table}

As demonstrated in \hyperref[tab:2]{Table 2}, we contrast our methodology, VCP, with the T5 baseline and other methods using the ViGGO dataset. Our method exhibits a notable advantage in terms of reducing the Slot Error Rate (SER) while maintaining comparable non-SER scores to the T5 baseline. The SER result reported by K\&M cannot be directly compared with our method, given that we employ different methodologies for calculating SER. Our VCP method reduces SER from 0.89\% to 0.41\% on T5-small and from 0.60\% to 0.33\% on T5-base.

As shown in \hyperref[tab:3]{Table 3}, for the E2E dataset, our VCP method not only retains text generation quality on non-SER evaluation metrics but also reduces SER from over 2.5\% for the T5 baseline to almost 0. This is lower than other methods, except for K\&M. Although K\&M performs well on the E2E dataset, it struggles to maintain text generation quality, achieving a 48.5 BLEU score on the ViGGO dataset. DM-NLG, with post-processing using GPT-2, reduces SER to 0.03\% and improves text fluency on non-SER evaluation metrics. However, their method incorporates a post-processing stage that employs a significantly larger language model, GPT-2, to enhance the fluency of the initial prediction. This makes the text quality comparison with our method somewhat unfair. Our method achieves a higher BLEU score compared to DM-NLG without post-processing, while reaching a lower SER score.

We also contrast our approach with T5 S\&L, a few-shot post-editing method. On the E2E dataset, they utilized the T5-base model, training it with 420 labeled and 1680 paralleled data points. Our approach involved training the T5-base model on the 420 labeled examples from the E2E dataset and then creating a training set for prompt generation from the first 2100 (420 labeled + 1680 unlabeled) data points of the E2E dataset. We adhered to the same hyperparameters as those outlined in our study when assessing the T5-small model on the E2E dataset. After three trials, we achieved an SER of 0.55\%, which was significantly better than their reported 0.8\%. Our BLEU score was 60.52, closely matching their score of 60.89.

\subsection{Comparing to LLMs}
\label{sec:GPT3.5}

\hyperref[tab:4]{Table 4} summarizes our evaluation of GPT3.5 and GPT4. It is critical to note that the performance of GPT3.5, when given five in-context examples of the input intent type (as shown in Appendix-Prompt with One Example for Each Intent), significantly underperforms compared to the T5 baseline in terms of the BLEU score. This underperformance is also evident when GPT3.5 is given one in-context example for every intent from the ViGGO dataset (details in Appendix-Prompt with Selected Examples). The same prompting strategy was applied to GPT4. Although GPT4 does not achieve a high BLEU score, it attains considerably lower SER scores. Remarkably, GPT4 achieved a 0\% SER with selected in-context examples, showcasing its exceptional capability to accurately follow instructions.

We believe the lower prediction quality generated by GPT3.5 and GPT4 is primarily due to the complexities involved in expressing the relationship between the input and output through in-context examples and prompts. For instance, despite being provided with five distinct request attribute examples and clear prompt explanations (as detailed in Appendix-Prompt with Selected Examples), GPT3.5 falls short in accurately replicating the desired tone and often misconstrues the intended meaning of the request attribute intent in the input data. For example, in the data shown in Appendix-Prompt with Selected Examples, the intent of the request attribute suggests that the user is seeking to ascertain whether their feelings are average. GPT3.5 misinterprets this, inferring that the input data is attempting to verify all available information. There are myriad ways to interpret how an AI model should convert input data into text. However, the true relationship can be more effectively understood through training a language model on thousands of examples, rather than presenting it with a limited number of in-context examples and descriptions. Consequently, supervised training continues to play an essential role in data-to-text generation models.

\section{Ablation Study}

\begin{table}[t]
\caption{Performance comparison between LLMs, T5 baseline and our method}
\vspace{-0.3cm}
\centering
\begin{tabular}{lll}
\hline
                & \multicolumn{1}{c}{BLEU} & \multicolumn{1}{c}{SER} \\\hline
T5-small$_{beam search}$ baseline
   & 53.2           & 0.89\%                   \\
VCP$_{T5small}$  & 52.6                   & \textbf{0.41\%}            
\\ \hline
GPT3.5 examples       &         25.1               &          7.3\%          \\
GPT3.5 selected examples   & 22.3                       & 6.72\%                                             \\

GPT4 examples       &         30.7            &        0.90\%        \\
GPT4 selected examples   & 27.4                      &  \textbf{0}                                           \\

\hline
\end{tabular}%
\label{tab:4}
\vspace{-0.3cm}
\end{table}

In \hyperref[tab:5]{Table 5}, we conduct an ablation study using the T5-small model on the ViGGO dataset.

\begin{table}[t]
\caption{Ablation study on ViGGO dataset}
\vspace{-0.3cm}
\centering
\begin{tabular}{lllllll}
\hline
                       & BLEU      &  SER   \\ \hline
T5-small$_{beam search}$ baseline            & 53.2     &  0.89\% \\
VCP$_{T5small}$           & 52.6      &  0.41\%       \\\hline
Remove position information       &  52.4     &  0.72\%      \\
Fine-tune T5(not use prompt)             & 51.4     &  0.70\%      \\
No prompt initilization          &  52.6     &  0.65\%    \\
Directly sampling the output              &  50.6     &  0.62\%     \\ \hline
\end{tabular}
\label{tab:5}
\vspace{-0.3cm}
\end{table}

\subsection{Removing Position Information}
In this experiment, we evaluate the importance of placing error-correcting prompts adjacent to the slot errors. To this end, we remove the slot error position information by always positioning the error-correcting prompts at the front of the inputs. \hyperref[tab:5]{Table 5} illustrates that slot error rates rise when we remove information regarding error locations, underscoring the advantage of highlighting the probable sites of errors.

\subsection{Fine-tuning the Entire Model (T5 no prompt)}
In this experiment, we assess whether using error-correcting prompts results in better performance than fine-tuning the entire model on the generated training data. After training a T5 model on the training dataset, we further fine-tune it with a learning rate of 5\^e-3 for 10 epochs using the prompt training dataset (with error-correcting prompts removed) created in the Data Generation section. We compare its performance to VCP, which only trains error-correcting prompts. As shown in \hyperref[tab:5]{Table 5}, the BLEU score and SER generated by 'Fine-tune T5' are noticeably lower than those achieved by prompt tuning methods, demonstrating the importance of using prompts. The reason lies in the fact that prompts not only label the error locations but also allow the original model to remain frozen. The original model, trained on ground-truth data labeled by humans (unlike ground-truth in the prompt training datasets created by T5 models which may contain errors), excels at producing high-quality texts. Utilizing prompts enables the minimally affected well-trained T5 model, thereby yielding better-quality text outputs (high BLEU score) and learning a more generalizable ability to guide language models in reducing slot errors.

\subsection{Remove the prompt initialization training process}
\label{sec:promptinitialization}
During the training process, we first train prompt initialization and then fine-tune the initialization embedding, as opposed to directly fine-tuning a randomly initialized embedding. In this experiment, we aim to evaluate the significance of prompt initialization by comparing the performance of VCP before and after using the prompt initialization step (Not initial.). As shown in \hyperref[tab:5]{Table 5}, there is a decline in the BLEU score and an increase in the SER score after the removal of prompt initialization, thereby emphasizing its vital role in maintaining the quality of text generation.

\subsection{Sampling the best prediction directly}
In our project, we use error-correcting prompts to guide fine-tuned T5 models in correcting their slot errors. We compare our method to directly sampling 10 outputs and selecting the best prediction using an SER score. The results demonstrate that while the direct sampling method reduces the SER, the quality of the generated texts diminishes compared to the prompt-based method. The main reason for this is that the slot error checker does not always accurately recognize when generated texts use different words to mention slot information. This can result in the original prediction being incorrectly identified as having errors, leading to the selection of alternative predictions which may have lower text fluency. Error-correcting prompts, on the other hand, can learn generalizable knowledge during training, allowing them to guide the T5 model beyond the sampling search range and perform more natural predictions.

\section{Limitations and Future Work}

Our method effectively reduces slot errors; however, it also slightly decreases text fluency. This happens because we train the prompt tokens using the ground truth generated by the fine-tuned language model itself, which can sometimes be inaccurate or sound unnatural. The introduction of a filter to eliminate low-quality text, or the use of post-processing tools to enhance the quality of the generated text, may improve text fluency.

Another limitation is that the knowledge imparted by the prompt training datasets is limited to the capabilities of the posterior checker (such as our slot error checker). If we cannot check for certain types of errors, then we are unable to construct the prompt training datasets that impart that specific kind of knowledge. This issue could potentially be resolved by training specific error-detection models or using more powerful language models that can generally check most error types. In this way, we would not need to rely on hard-to-design rules when the knowledge we are trying to impart is complex.

\section{Discussion and Conclusion}
By utilizing a feedback system pipeline, our method achieves the lowest SER compared to other methods, while still maintaining a comparable level of text generation quality. Our approach attains a lower SER and maintains text quality primarily because it does not overly rely on predefined rules, which can be inaccurate in complex scenarios. Our specialized training method enables accurate regeneration even with imprecise feedback. Additionally, our feedback system not only informs the model about the correctness of its output but also indicates where the errors are located.

Our method is particularly suitable for smaller models that are incapable of reasoning based on natural language feedback. To preserve the efficiency advantage of such models, we use a basic verifier. While this verifier is not highly accurate, it is both easy to implement and fast, making it an efficient choice.

We hope VCP can adapt to various applications requiring feedback systems, especially in cases where providing or interpreting feedback is challenging. This encompasses applications such as text-to-image generation, story summarization, and text-to-SQL generation.

\clearpage

\clearpage
\appendix
\section{Appendix}
\label{sec:appendix}

\subsection{Detailed Explanation of the Slot Error Checking Process}
\label{sec:slot_error_checker_explaination}
We employ a straightforward implementation of the slot error checker. We use the same slot error checker for ViGGO and E2E datasets. This checker do not contain any domain spefic knowledge and it follows a simple rule: it verifies whether the slot value is mentioned in the prediction through exact match. For instance, if the input slot is "price: [high]" and the output fails to mention "high," the slot error checker will identify this as an error and append a special token next to "high." When dealing with a boolean slot value (yes or no), such as "has\_linux\_released: [no]," the checker isolates the noun component 'linux' and checks for its presence in the prediction. Its absence triggers the checker to mark an error-correcting soft prompt next to 'no'. It's important to note that our approach does not presuppose the infallibility of the slot error checker. For example, with an input slot of "price: [high]" and an output stating "it is expensive," the checker would still flag a slot error. However, the error-correcting prompt in the regeneration process will ignore such incorrect labeling, ensuring the generation of accurate output nonetheless, which means the error-correcting prompts are robust to the mislabel senarios. More details can be refer to Appendix~\hyperref[sec:error_correcting_prompts_robustness]{Robustness to Slot Checking Errors}.

A detailed bollean example: If the input is \textit{has\_linux\_released[yes]}, the first step is to see if \textit{linux} appears in the prediction. If it does not, error-correcting prompts are positioned beside the slot value [yes], change it to [<token1><token2><token3>yes] if we use 3 trainable prompts. If it does appear, we employ simple dependency parsing rules and POS tags to ascertain if any negation words are linked to \textit{linux}. If negation words are found, the slot value is marked with a error-correcting prompts. If no negation words are present, we infer that there are no slot errors concerning \textit{has\_linux\_released[yes]}. Conversely, if the slot value is \textit{no}, such as in \textit{has\_linux\_released[no]}, the initial step is to check for the mention of \textit{linux} in the prediction. If \textit{linux} is not mentioned, it is assumed that no slot errors exist. However, if \textit{linux} is mentioned, we look for any associated negation words. If none are found, error-correcting prompts are placed beside the \textit{no} slot value, indicating a potential slot error. If negation words are present, we presume the absence of slot errors.

\subsection{Robustness to Slot Checking Errors}
\label{sec:error_correcting_prompts_robustness}

The error-correcting prompt exhibits considerable resilience in the face of inaccurately labeled data across both training and evaluation datasets. The slot error checker frequently errs in identifying inaccuracies. Nonetheless, the fidelity of the outputs generated remains elevated, owing to the error-correcting prompts’ proficient management of inputs with erroneous labels. Consider, for instance, an input might contains the slot 'rating = average', juxtaposed with an output declaration that 'Black Hole Entertainment always makes games that are at least okay'. The slot error checker that rely on simple rules neglects to recognize the implication of 'at least ok' as equivalent to an 'average' rating. 

In conducting experiments with the T5-base model on the ViGGO dataset, the slot error checker erroneously suggests an average of 128.5 errors. Our empirical findings indicate a slot error rate of 0.60\% for the T5-base model. Given the dataset comprises 359 testing instances with a total of 1372 slot values, a 0.60\% error rate corresponds to roughly 8.2 actual slot errors. This analysis reveals that over 90\% of the errors labeled by the slot error checker are false positives.

On E2E dataset with VCP method applied to T5-base model, the naive slot error checker highlighting an average of 35.5 errors per prediction. Based on our data, the real slot error rate is adjusted to 0.33\%, suggesting a true error count of approximately 4.5. This discrepancy underscores the inaccuracy of the slot error checker.

The data demonstrates that error-correcting prompts don't always require including potential errors in every scenario. Many predictions lack the slot values identified by slot error checker after applying the error-correcting prompt, yet still, most outcomes are correct. This implies that the error correcting prompt does not always enforce the model to include the labeled slot values. Otherwise we would expect to see significantly fewer errors detected by the slot error checker. This indicates that error-correcting prompts are generally robust against mislabeling. Such robustness allows for the use of simple, less accurate rule-based slot checkers in real-world machine learning implementations.

\subsection{Prompt with Selected Examples}
We demonstrate an example of the prompt we use for GPT3.5. We demonstrate the prompt for random select the example from training dataset with the intent that is the same as the intent in the test example (request attribute). 

\begin{spverbatim}
PROMPT:


please perform data-to-text generation for me. Domain is video game. The words before the bracket are intentions.

For example, when the intention is give opinion, then the output should be a sentence that asks for opinion.

when the intention is verify attribute, then the output should be a sentence that try to verify the attribute.

Example:
Input: request attribute(esrb[])
Output: Are there any ESRB content ratings which you give preference to when picking a game to play?

Input: request attribute(release year[])
Output: Can you think of a year, in which video games were particularly good?

Input: request attribute(esrb[])
Output: Are there any ESRB content ratings which you give preference to when picking a game to play?

Input: request attribute(esrb[])
Output: Are there any ESRB content ratings which you give preference to when picking a game to play?

Input: request attribute(developer[])
Output: Which game developer do you think is the best?

Question:
Input: verify attribute(name[little big adventure], 
rating[average], has multiplayer[no], 
platforms[playstation])

Output: 


Answer by GPT3.5: Can you confirm that Little Big Adventure has an average rating and does not have multiplayer? Also, is it available on PlayStation?

GroundTruth:
['I remember you saying you found Little Big Adventure to be average. Are you not usually that into single-player games on PlayStation?', "Earlier, you stated that you didn't have strong feelings about PlayStation's Little Big Adventure. Is your opinion true for all games which don't have multiplayer?", 'I recall that you were not that fond of Little Big Adventure. Does single-player gaming on the PlayStation quickly get boring for you?']
\end{spverbatim}

\subsection{Prompt with One Example for Each Intent}
We demonstrate an example of the prompt we use for GPT3.5. In the example in the Appendix-Prompt with One Example for Each Intent, we demonstrate the prompt we use for all of the test cases. The prompt include one example for each intent.

\begin{spverbatim}
PROMPT:

please perform data-to-text generation for me. Domain is video game. 
the words before the bracket are intentions. For example, when the intention is give opinion, then the output should be a sentence that asks for opinion.
when the intention is verify attribute, then the output should be a sentence that try to verify the attribute. 

Example for each intention:
Input: give opinion(NAME [SpellForce 3], RATING [poor], GENRES [real-time strategy, role-playing], 
PLAYER PERSPECTIVE [bird view])
Output: I think that SpellForce 3 is one of the worst games I’ve ever played. Trying to combine the real-time strategy and role-playing genres just doesn’t work, and the bird’s eye view makes it near impossible to play.

Input: verify attribute(NAME [Little Big Adventure], RATING [average], HAS MULTIPLAYER [no], 
PLATFORMS [PlayStation])
Output: I recall that you were not that fond of Little Big Adventure. Does single-player gaming on the PlayStation quickly get boring for you?

Input: confirm(NAME [Hellblade: Senua’s Sacrifice], RELEASE YEAR [2017], DEVELOPER [Ninja Theory])
Output: Oh, do you mean the 2017 game from Ninja Theory, Hellblade: Senua’s Sacrifice?

Input: request(SPECIFIER [interesting])
Output: Have you played any interesting games lately?

Input: suggest(NAME [Half-Life 2], GENRES [shooter], PLAYER PERSPECTIVE [first person])
Output: Do you also enjoy playing first-person shooters, such as Half-Life 2?

Input: request explanation(RATING [poor], HAS MAC RELEASE [yes])
Output: What is it about Mac games that you find so disappointing?

Input: inform(NAME [Max Payne 3], RELEASE YEAR [2012], GENRES [action-adventure; shooter], MULTIPLAYER [yes])
Output: Max Payne 3 is a multiplayer action-adventure shooter from 2012.

Input: request attribute(AVAILABLE ON STEAM [])
Output: Do you prefer playing games that you can get on Steam?

Question:
Input: YOUR INPUT QUESTION
Output: 

\end{spverbatim}

\subsection{Training parameters}
We use the learning rate begins at 0.01 and reduces gradually over 20 epochs for prompt embedding initialization. We use 3 error-correcting prompts for T5-base and 6 error-correcting prompts for T5-small. More details can be seen in Table 6 and 7.

\subsection{Experiment details}
On the ViGGO dataset, we run the T5 baseline, SEA-GUIDE project, and our VCP for 5 times. We report the mean and variance in \hyperref[tab:2]{Table 2} and reproduce the experiment results of S2S, DT, and K\&M from the SEA-GUIDE paper. We also run the ablation study once and report the results in Tables 4 and 5.

For the E2E dataset, we run our VCP and reported the mean and variance. We report the experiment results from the DM-NLG paper and the SEA-GUIDE paper in \hyperref[tab:3]{Table 3}. 

We use the SER auto-evaluation script from the SEA-GUIDE Github project to evaluate SER on both the ViGGO and E2E datasets. However, when evaluating the model on the E2E dataset, the SER evaluation script is not accurate. As a result, we manually check every prediction labeled as having slot errors by the SER evaluation script. When applying GPT3.5 and GPT4 to the ViGGO dataset, we also manually check every prediction labeled as incorrect by the SER evaluation script.

\subsection{Prompt Initialization Details}
To achieve this, we train error-correction prompts such that inserting them does not alter the output of the fine-tuned T5 model. Specifically, we remove the prompts from the unseen prompted input, then forward these inputs to the fine-tuned T5 model for prediction. We use these predictions as our training targets and the unseen prompted inputs as training input. We train the error correction prompt tokens while remain fine-tuned T5 model frozen. Performing prompt tuning on such a initialized prompt instead of the random initialized prompt is demonstrated to have better performance as shown in the prompt initialization ablation study.

\subsection{Adding Extra Instruction for GPT}
We suspect that the reason why ChatGPT may not always follow instructions perfectly is because we have not specifically encouraged it to do so. Consequently, we conducted the "ChatGPT with Selected In-Context Examples" experiment three times, incorporating an additional instruction: "Please also ensure that all of the keywords in the inputs are mentioned." The results yielded a 7.96\% Slot Error Rate (SER) and a 22.51 BLEU score. Interestingly, the extra instruction did not enhance either the BLEU or SER scores, likely because ChatGPT already understood what was required by examining the five in-context examples.

\begin{table}
\caption{The following details pertain to the training process of our VCP method for experiments on the ViGGO datasets. 'Initial. lr' stands for the initial learning rate used for prompt initialization, while 'Train. lr' represents the learning rate used for prompt tuning. 'Epochs' refers to the number of epochs for prompt tuning.}
\centering
\begin{tabular}{lllll}
\hline
Model          & initial. lr & train. lr & epochs & prompt token num  \\ \hline
VCP$_{T5-small}$ & 0.01            &  0.005         & 5  &  6      \\
VCP$_{T5-base}$  & 0.01            &  0.01         & 2  &  3     \\ \hline
\end{tabular}
\end{table}

\begin{table}
\caption{The following details pertain to the training process of our VCP method for experiments on the E2E datasets. 'Initial. lr' stands for the initial learning rate used for prompt initialization, while 'Train. lr' represents the learning rate used for prompt tuning. 'Epochs' refers to the number of epochs for prompt tuning.}
\centering
\begin{tabular}{lllll}

\hline
Model          & initial. lr & train. lr & epochs & prompt token num \\ \hline
VCP$_{T5-small}$  & 0.01            & 0.01          & 10   & 6    \\
\hline
\end{tabular}
\end{table}

\begin{thebibliography}{27}


\ifx \showCODEN    \undefined \def \showCODEN     #1{\unskip}     \fi
\ifx \showISBNx    \undefined \def \showISBNx     #1{\unskip}     \fi
\ifx \showISBNxiii \undefined \def \showISBNxiii  #1{\unskip}     \fi
\ifx \showISSN     \undefined \def \showISSN      #1{\unskip}     \fi
\ifx \showLCCN     \undefined \def \showLCCN      #1{\unskip}     \fi
\ifx \shownote     \undefined \def \shownote      #1{#1}          \fi
\ifx \showarticletitle \undefined \def \showarticletitle #1{#1}   \fi
\ifx \showURL      \undefined \def \showURL       {\relax}        \fi
\providecommand\bibfield[2]{#2}
\providecommand\bibinfo[2]{#2}
\providecommand\natexlab[1]{#1}
\providecommand\showeprint[2][]{arXiv:#2}

\bibitem[Balachandran et~al\mbox{.}(2022)]%
        {balachandran-etal-2022-correcting}
\bibfield{author}{\bibinfo{person}{Vidhisha Balachandran}, \bibinfo{person}{Hannaneh Hajishirzi}, \bibinfo{person}{William Cohen}, {and} \bibinfo{person}{Yulia Tsvetkov}.} \bibinfo{year}{2022}\natexlab{}.
\newblock \showarticletitle{Correcting Diverse Factual Errors in Abstractive Summarization via Post-Editing and Language Model Infilling}. In \bibinfo{booktitle}{\emph{Proceedings of the 2022 Conference on Empirical Methods in Natural Language Processing}}, \bibfield{editor}{\bibinfo{person}{Yoav Goldberg}, \bibinfo{person}{Zornitsa Kozareva}, {and} \bibinfo{person}{Yue Zhang}} (Eds.). \bibinfo{publisher}{Association for Computational Linguistics}, \bibinfo{address}{Abu Dhabi, United Arab Emirates}, \bibinfo{pages}{9818--9830}.
\newblock
\href{https://doi.org/10.18653/v1/2022.emnlp-main.667}{doi:\nolinkurl{10.18653/v1/2022.emnlp-main.667}}


\bibitem[Harkous et~al\mbox{.}(2020)]%
        {harkous-etal-2020-text}
\bibfield{author}{\bibinfo{person}{Hamza Harkous}, \bibinfo{person}{Isabel Groves}, {and} \bibinfo{person}{Amir Saffari}.} \bibinfo{year}{2020}\natexlab{}.
\newblock \showarticletitle{Have Your Text and Use It Too! End-to-End Neural Data-to-Text Generation with Semantic Fidelity}. In \bibinfo{booktitle}{\emph{Proceedings of the 28th International Conference on Computational Linguistics}}. \bibinfo{publisher}{International Committee on Computational Linguistics}, \bibinfo{address}{Barcelona, Spain (Online)}.
\newblock
\href{https://doi.org/10.18653/v1/2020.coling-main.218}{doi:\nolinkurl{10.18653/v1/2020.coling-main.218}}


\bibitem[Jolly et~al\mbox{.}(2022)]%
        {jolly2022search}
\bibfield{author}{\bibinfo{person}{Shailza Jolly}, \bibinfo{person}{Zi~Xuan Zhang}, \bibinfo{person}{Andreas Dengel}, {and} \bibinfo{person}{Lili Mou}.} \bibinfo{year}{2022}\natexlab{}.
\newblock \showarticletitle{Search and learn: improving semantic coverage for data-to-text generation}. In \bibinfo{booktitle}{\emph{Proceedings of the AAAI Conference on Artificial Intelligence}}, Vol.~\bibinfo{volume}{36}. \bibinfo{pages}{10858--10866}.
\newblock


\bibitem[Juraska et~al\mbox{.}(2019)]%
        {juraska-etal-2019-viggo}
\bibfield{author}{\bibinfo{person}{Juraj Juraska}, \bibinfo{person}{Kevin Bowden}, {and} \bibinfo{person}{Marilyn Walker}.} \bibinfo{year}{2019}\natexlab{}.
\newblock \showarticletitle{{V}i{GGO}: A Video Game Corpus for Data-To-Text Generation in Open-Domain Conversation}. In \bibinfo{booktitle}{\emph{Proceedings of the 12th International Conference on Natural Language Generation}}. \bibinfo{publisher}{Association for Computational Linguistics}, \bibinfo{address}{Tokyo, Japan}.
\newblock
\href{https://doi.org/10.18653/v1/W19-8623}{doi:\nolinkurl{10.18653/v1/W19-8623}}


\bibitem[Juraska and Walker(2021)]%
        {juraska-walker-2021-attention}
\bibfield{author}{\bibinfo{person}{Juraj Juraska} {and} \bibinfo{person}{Marilyn Walker}.} \bibinfo{year}{2021}\natexlab{}.
\newblock \showarticletitle{Attention Is Indeed All You Need: Semantically Attention-Guided Decoding for Data-to-Text {NLG}}. In \bibinfo{booktitle}{\emph{Proceedings of the 14th International Conference on Natural Language Generation}}. \bibinfo{publisher}{Association for Computational Linguistics}, \bibinfo{address}{Aberdeen, Scotland, UK}, \bibinfo{pages}{416--431}.
\newblock
\urldef\tempurl%
\url{https://aclanthology.org/2021.inlg-1.45}
\showURL{%
\tempurl}


\bibitem[Kale and Rastogi(2020)]%
        {kale2020template}
\bibfield{author}{\bibinfo{person}{Mihir Kale} {and} \bibinfo{person}{Abhinav Rastogi}.} \bibinfo{year}{2020}\natexlab{}.
\newblock \showarticletitle{Template guided text generation for task-oriented dialogue}.
\newblock \bibinfo{journal}{\emph{arXiv preprint arXiv:2004.15006}} (\bibinfo{year}{2020}).
\newblock


\bibitem[Kasner and Dusek(2022)]%
        {kasner-dusek-2022-neural}
\bibfield{author}{\bibinfo{person}{Zden{\v{e}}k Kasner} {and} \bibinfo{person}{Ondrej Dusek}.} \bibinfo{year}{2022}\natexlab{}.
\newblock \showarticletitle{Neural Pipeline for Zero-Shot Data-to-Text Generation}. In \bibinfo{booktitle}{\emph{Proceedings of the 60th Annual Meeting of the Association for Computational Linguistics (Volume 1: Long Papers)}}. \bibinfo{publisher}{Association for Computational Linguistics}, \bibinfo{address}{Dublin, Ireland}.
\newblock
\urldef\tempurl%
\url{https://aclanthology.org/2022.acl-long.271}
\showURL{%
\tempurl}


\bibitem[Kedzie and McKeown(2020)]%
        {kedzie-mckeown-2020-controllable}
\bibfield{author}{\bibinfo{person}{Chris Kedzie} {and} \bibinfo{person}{Kathleen McKeown}.} \bibinfo{year}{2020}\natexlab{}.
\newblock \showarticletitle{Controllable Meaning Representation to Text Generation: Linearization and Data Augmentation Strategies}. In \bibinfo{booktitle}{\emph{Proceedings of the 2020 Conference on Empirical Methods in Natural Language Processing (EMNLP)}}. \bibinfo{publisher}{Association for Computational Linguistics}, \bibinfo{address}{Online}, \bibinfo{pages}{5160--5185}.
\newblock
\href{https://doi.org/10.18653/v1/2020.emnlp-main.419}{doi:\nolinkurl{10.18653/v1/2020.emnlp-main.419}}


\bibitem[Lavie and Agarwal(2007)]%
        {lavie-agarwal-2007-meteor}
\bibfield{author}{\bibinfo{person}{Alon Lavie} {and} \bibinfo{person}{Abhaya Agarwal}.} \bibinfo{year}{2007}\natexlab{}.
\newblock \showarticletitle{{METEOR}: An Automatic Metric for {MT} Evaluation with High Levels of Correlation with Human Judgments}. In \bibinfo{booktitle}{\emph{Proceedings of the Second Workshop on Statistical Machine Translation}}. \bibinfo{publisher}{Association for Computational Linguistics}, \bibinfo{address}{Prague, Czech Republic}, \bibinfo{pages}{228--231}.
\newblock
\urldef\tempurl%
\url{https://aclanthology.org/W07-0734}
\showURL{%
\tempurl}


\bibitem[Lin(2004)]%
        {lin-2004-rouge}
\bibfield{author}{\bibinfo{person}{Chin-Yew Lin}.} \bibinfo{year}{2004}\natexlab{}.
\newblock \showarticletitle{{ROUGE}: A Package for Automatic Evaluation of Summaries}. In \bibinfo{booktitle}{\emph{Text Summarization Branches Out}}. \bibinfo{publisher}{Association for Computational Linguistics}, \bibinfo{address}{Barcelona, Spain}, \bibinfo{pages}{74--81}.
\newblock
\urldef\tempurl%
\url{https://aclanthology.org/W04-1013}
\showURL{%
\tempurl}


\bibitem[Liu et~al\mbox{.}(2022)]%
        {liu-etal-2022-p}
\bibfield{author}{\bibinfo{person}{Xiao Liu}, \bibinfo{person}{Kaixuan Ji}, \bibinfo{person}{Yicheng Fu}, \bibinfo{person}{Weng Tam}, \bibinfo{person}{Zhengxiao Du}, \bibinfo{person}{Zhilin Yang}, {and} \bibinfo{person}{Jie Tang}.} \bibinfo{year}{2022}\natexlab{}.
\newblock \showarticletitle{{P}-Tuning: Prompt Tuning Can Be Comparable to Fine-tuning Across Scales and Tasks}. In \bibinfo{booktitle}{\emph{Proceedings of the 60th Annual Meeting of the Association for Computational Linguistics (Volume 2: Short Papers)}}. \bibinfo{publisher}{Association for Computational Linguistics}, \bibinfo{address}{Dublin, Ireland}, \bibinfo{pages}{61--68}.
\newblock
\href{https://doi.org/10.18653/v1/2022.acl-short.8}{doi:\nolinkurl{10.18653/v1/2022.acl-short.8}}


\bibitem[Madaan et~al\mbox{.}(2023)]%
        {madaan2023self}
\bibfield{author}{\bibinfo{person}{Aman Madaan}, \bibinfo{person}{Niket Tandon}, \bibinfo{person}{Prakhar Gupta}, \bibinfo{person}{Skyler Hallinan}, \bibinfo{person}{Luyu Gao}, \bibinfo{person}{Sarah Wiegreffe}, \bibinfo{person}{Uri Alon}, \bibinfo{person}{Nouha Dziri}, \bibinfo{person}{Shrimai Prabhumoye}, \bibinfo{person}{Yiming Yang}, {et~al\mbox{.}}} \bibinfo{year}{2023}\natexlab{}.
\newblock \showarticletitle{Self-refine: Iterative refinement with self-feedback}.
\newblock \bibinfo{journal}{\emph{arXiv preprint arXiv:2303.17651}} (\bibinfo{year}{2023}).
\newblock


\bibitem[Mehta et~al\mbox{.}(2022)]%
        {mehta-etal-2022-improving}
\bibfield{author}{\bibinfo{person}{Sanket~Vaibhav Mehta}, \bibinfo{person}{Jinfeng Rao}, \bibinfo{person}{Yi Tay}, \bibinfo{person}{Mihir Kale}, \bibinfo{person}{Ankur Parikh}, {and} \bibinfo{person}{Emma Strubell}.} \bibinfo{year}{2022}\natexlab{}.
\newblock \showarticletitle{Improving Compositional Generalization with Self-Training for Data-to-Text Generation}. In \bibinfo{booktitle}{\emph{Proceedings of the 60th Annual Meeting of the Association for Computational Linguistics (Volume 1: Long Papers)}}. \bibinfo{publisher}{Association for Computational Linguistics}, \bibinfo{address}{Dublin, Ireland}.
\newblock
\urldef\tempurl%
\url{https://aclanthology.org/2022.acl-long.289}
\showURL{%
\tempurl}


\bibitem[Novikova et~al\mbox{.}(2017)]%
        {novikova-etal-2017-e2e}
\bibfield{author}{\bibinfo{person}{Jekaterina Novikova}, \bibinfo{person}{Ond{\v{r}}ej Du{\v{s}}ek}, {and} \bibinfo{person}{Verena Rieser}.} \bibinfo{year}{2017}\natexlab{}.
\newblock \showarticletitle{The {E}2{E} Dataset: New Challenges For End-to-End Generation}. In \bibinfo{booktitle}{\emph{Proceedings of the 18th Annual {SIG}dial Meeting on Discourse and Dialogue}}. \bibinfo{publisher}{Association for Computational Linguistics}, \bibinfo{address}{Saarbr{\"u}cken, Germany}, \bibinfo{pages}{201--206}.
\newblock
\href{https://doi.org/10.18653/v1/W17-5525}{doi:\nolinkurl{10.18653/v1/W17-5525}}


\bibitem[Papineni et~al\mbox{.}(2002)]%
        {papineni2002bleu}
\bibfield{author}{\bibinfo{person}{Kishore Papineni}, \bibinfo{person}{Salim Roukos}, \bibinfo{person}{Todd Ward}, {and} \bibinfo{person}{Wei-Jing Zhu}.} \bibinfo{year}{2002}\natexlab{}.
\newblock \showarticletitle{Bleu: a method for automatic evaluation of machine translation}. In \bibinfo{booktitle}{\emph{Proceedings of the 40th annual meeting of the Association for Computational Linguistics}}. \bibinfo{pages}{311--318}.
\newblock


\bibitem[Peng et~al\mbox{.}(2023)]%
        {peng2023check}
\bibfield{author}{\bibinfo{person}{Baolin Peng}, \bibinfo{person}{Michel Galley}, \bibinfo{person}{Pengcheng He}, \bibinfo{person}{Hao Cheng}, \bibinfo{person}{Yujia Xie}, \bibinfo{person}{Yu Hu}, \bibinfo{person}{Qiuyuan Huang}, \bibinfo{person}{Lars Liden}, \bibinfo{person}{Zhou Yu}, \bibinfo{person}{Weizhu Chen}, {et~al\mbox{.}}} \bibinfo{year}{2023}\natexlab{}.
\newblock \showarticletitle{Check your facts and try again: Improving large language models with external knowledge and automated feedback}.
\newblock \bibinfo{journal}{\emph{arXiv preprint arXiv:2302.12813}} (\bibinfo{year}{2023}).
\newblock


\bibitem[Puduppully et~al\mbox{.}(2019)]%
        {puduppully2019data}
\bibfield{author}{\bibinfo{person}{Ratish Puduppully}, \bibinfo{person}{Li Dong}, {and} \bibinfo{person}{Mirella Lapata}.} \bibinfo{year}{2019}\natexlab{}.
\newblock \showarticletitle{Data-to-text generation with content selection and planning}. In \bibinfo{booktitle}{\emph{Proceedings of the AAAI conference on artificial intelligence}}, Vol.~\bibinfo{volume}{33}. \bibinfo{pages}{6908--6915}.
\newblock


\bibitem[Raffel et~al\mbox{.}(2020)]%
        {10.5555/3455716.3455856}
\bibfield{author}{\bibinfo{person}{Colin Raffel}, \bibinfo{person}{Noam Shazeer}, \bibinfo{person}{Adam Roberts}, \bibinfo{person}{Katherine Lee}, \bibinfo{person}{Sharan Narang}, \bibinfo{person}{Michael Matena}, \bibinfo{person}{Yanqi Zhou}, \bibinfo{person}{Wei Li}, {and} \bibinfo{person}{Peter~J. Liu}.} \bibinfo{year}{2020}\natexlab{}.
\newblock \showarticletitle{Exploring the Limits of Transfer Learning with a Unified Text-to-Text Transformer}.
\newblock \bibinfo{journal}{\emph{J. Mach. Learn. Res.}} \bibinfo{volume}{21}, \bibinfo{number}{1}, Article \bibinfo{articleno}{140} (\bibinfo{date}{jan} \bibinfo{year}{2020}), \bibinfo{numpages}{67}~pages.
\newblock
\showISSN{1532-4435}


\bibitem[Rebuffel et~al\mbox{.}(2019)]%
        {rebuffel2019hierarchical}
\bibfield{author}{\bibinfo{person}{Clément Rebuffel}, \bibinfo{person}{Laure Soulier}, \bibinfo{person}{Geoffrey Scoutheeten}, {and} \bibinfo{person}{Patrick Gallinari}.} \bibinfo{year}{2019}\natexlab{}.
\newblock \bibinfo{title}{A Hierarchical Model for Data-to-Text Generation}.
\newblock
\showeprint[arxiv]{1912.10011}~[cs.CL]


\bibitem[Seifossadat and Sameti(2023)]%
        {seifossadat2023improving}
\bibfield{author}{\bibinfo{person}{Elham Seifossadat} {and} \bibinfo{person}{Hossein Sameti}.} \bibinfo{year}{2023}\natexlab{}.
\newblock \showarticletitle{Improving semantic coverage of data-to-text generation model using dynamic memory networks}.
\newblock \bibinfo{journal}{\emph{Natural Language Engineering}} (\bibinfo{year}{2023}), \bibinfo{pages}{1--26}.
\newblock


\bibitem[Shridhar et~al\mbox{.}(2023a)]%
        {shridhar2023screws}
\bibfield{author}{\bibinfo{person}{Kumar Shridhar}, \bibinfo{person}{Harsh Jhamtani}, \bibinfo{person}{Hao Fang}, \bibinfo{person}{Benjamin Van~Durme}, \bibinfo{person}{Jason Eisner}, {and} \bibinfo{person}{Patrick Xia}.} \bibinfo{year}{2023}\natexlab{a}.
\newblock \showarticletitle{SCREWS: A Modular Framework for Reasoning with Revisions}.
\newblock \bibinfo{journal}{\emph{arXiv preprint arXiv:2309.13075}} (\bibinfo{year}{2023}).
\newblock


\bibitem[Shridhar et~al\mbox{.}(2023b)]%
        {shridhar2023art}
\bibfield{author}{\bibinfo{person}{Kumar Shridhar}, \bibinfo{person}{Koustuv Sinha}, \bibinfo{person}{Andrew Cohen}, \bibinfo{person}{Tianlu Wang}, \bibinfo{person}{Ping Yu}, \bibinfo{person}{Ram Pasunuru}, \bibinfo{person}{Mrinmaya Sachan}, \bibinfo{person}{Jason Weston}, {and} \bibinfo{person}{Asli Celikyilmaz}.} \bibinfo{year}{2023}\natexlab{b}.
\newblock \showarticletitle{The ART of LLM Refinement: Ask, Refine, and Trust}.
\newblock \bibinfo{journal}{\emph{arXiv preprint arXiv:2311.07961}} (\bibinfo{year}{2023}).
\newblock


\bibitem[Su et~al\mbox{.}(2021)]%
        {su2021plan}
\bibfield{author}{\bibinfo{person}{Yixuan Su}, \bibinfo{person}{David Vandyke}, \bibinfo{person}{Sihui Wang}, \bibinfo{person}{Yimai Fang}, {and} \bibinfo{person}{Nigel Collier}.} \bibinfo{year}{2021}\natexlab{}.
\newblock \showarticletitle{Plan-then-generate: Controlled data-to-text generation via planning}.
\newblock \bibinfo{journal}{\emph{arXiv preprint arXiv:2108.13740}} (\bibinfo{year}{2021}).
\newblock


\bibitem[Vedantam et~al\mbox{.}(2015)]%
        {vedantam2015cider}
\bibfield{author}{\bibinfo{person}{Ramakrishna Vedantam}, \bibinfo{person}{C.~Lawrence Zitnick}, {and} \bibinfo{person}{Devi Parikh}.} \bibinfo{year}{2015}\natexlab{}.
\newblock \bibinfo{title}{CIDEr: Consensus-based Image Description Evaluation}.
\newblock
\showeprint[arxiv]{1411.5726}~[cs.CV]


\bibitem[Xu et~al\mbox{.}(2021)]%
        {xu-etal-2021-agggen}
\bibfield{author}{\bibinfo{person}{Xinnuo Xu}, \bibinfo{person}{Ond{\v{r}}ej Du{\v{s}}ek}, \bibinfo{person}{Verena Rieser}, {and} \bibinfo{person}{Ioannis Konstas}.} \bibinfo{year}{2021}\natexlab{}.
\newblock \showarticletitle{{A}gg{G}en: Ordering and Aggregating while Generating}. In \bibinfo{booktitle}{\emph{Proceedings of the 59th Annual Meeting of the Association for Computational Linguistics and the 11th International Joint Conference on Natural Language Processing (Volume 1: Long Papers)}}. \bibinfo{publisher}{Association for Computational Linguistics}, \bibinfo{address}{Online}.
\newblock


\bibitem[Xue et~al\mbox{.}(2023)]%
        {xue2023rcot}
\bibfield{author}{\bibinfo{person}{Tianci Xue}, \bibinfo{person}{Ziqi Wang}, \bibinfo{person}{Zhenhailong Wang}, \bibinfo{person}{Chi Han}, \bibinfo{person}{Pengfei Yu}, {and} \bibinfo{person}{Heng Ji}.} \bibinfo{year}{2023}\natexlab{}.
\newblock \showarticletitle{RCOT: Detecting and Rectifying Factual Inconsistency in Reasoning by Reversing Chain-of-Thought}.
\newblock \bibinfo{journal}{\emph{arXiv preprint arXiv:2305.11499}} (\bibinfo{year}{2023}).
\newblock


\bibitem[Yin and Wan(2022)]%
        {yin-wan-2022-seq2seq}
\bibfield{author}{\bibinfo{person}{Xunjian Yin} {and} \bibinfo{person}{Xiaojun Wan}.} \bibinfo{year}{2022}\natexlab{}.
\newblock \showarticletitle{How Do {S}eq2{S}eq Models Perform on End-to-End Data-to-Text Generation?}. In \bibinfo{booktitle}{\emph{Proceedings of the 60th Annual Meeting of the Association for Computational Linguistics (Volume 1: Long Papers)}}, \bibfield{editor}{\bibinfo{person}{Smaranda Muresan}, \bibinfo{person}{Preslav Nakov}, {and} \bibinfo{person}{Aline Villavicencio}} (Eds.). \bibinfo{publisher}{Association for Computational Linguistics}, \bibinfo{address}{Dublin, Ireland}, \bibinfo{pages}{7701--7710}.
\newblock
\href{https://doi.org/10.18653/v1/2022.acl-long.531}{doi:\nolinkurl{10.18653/v1/2022.acl-long.531}}


\end{thebibliography}
\end{document}